\documentclass{article}
\usepackage{graphicx}		
\DeclareGraphicsExtensions{.pdf,.jpeg,.png}
\usepackage{color}
\usepackage{subfig}
\usepackage{amssymb, amsmath}                            
\usepackage{amsfonts}
\usepackage{mathtools}
\usepackage{pdfpages}
\usepackage{tabularx}
\usepackage{booktabs}
\usepackage{colortbl} 
\usepackage{xcolor} 
\usepackage{xfrac}
\usepackage{multirow}
\usepackage{tabularx}
\usepackage[linesnumbered,ruled]{algorithm2e}
\usepackage[numbers,square,sort&compress]{natbib}
\usepackage{authblk}

\usepackage[utf8]{inputenc}

\title{Objective Surgical Skills Assessment and Tool Localization: Results from the MICCAI 2021 SimSurgSkill Challenge}
\author[1]{Aneeq Zia}
\author[1]{Kiran Bhattacharyya}
\author[1]{Xi Liu}
\author[1]{Ziheng Wang}
\author[1]{Max Berniker}
\author[2]{Satoshi Kondo}
\author[3]{Emanuele Colleoni}
\author[3]{Dimitris Psychogyios}
\author[3]{Yueming Jin}
\author[4]{Jinfan Zhou}
\author[3]{Evangelos Mazomenos}
\author[6]{Lena Maier-Hein}
\author[3]{Danail Stoyanov}
\author[7]{Stefanie Speidel}
\author[1]{Anthony Jarc}
\affil[1]{Intuitive, Inc.}
\affil[2]{Muroran Institute of Technology}
\affil[3]{University College London}
\affil[4]{Shandong University}
\affil[6]{German Cancer Research Center (DKFZ)}
\affil[7]{National Center for Tumor Diseases (NCT)}
\date{}

\begin{document}

\maketitle

% abstract: what is problem, how do you address it, what conclusion/benefit?
% a lot of paragraphs are missing a conclusion
% what is the simulations formal name?
% are these team names or people's names?
% descriptions are in differing tenses, persons and formats

\begin{abstract}
Timely and effective feedback within surgical training plays a critical role in developing the skills required to perform safe and efficient surgery. Feedback from expert surgeons, while especially valuable in this regard, is challenging to acquire due to their typically busy schedules, and may be subject to biases. Formal assessment procedures like OSATS and GEARS attempt to provide objective measures of skill, but remain time-consuming. With advances in machine learning there is an opportunity for fast and objective automated feedback on technical skills. The SimSurgSkill 2021 challenge (hosted as a sub-challenge of EndoVis at MICCAI 2021) aimed to promote and foster work in this endeavor. Using virtual reality (VR) surgical tasks, competitors were tasked with localizing instruments and predicting surgical skill. Here we summarize the winning approaches and how they performed. Using this publicly available dataset and results as a springboard, future work may enable more efficient training of surgeons with advances in surgical data science. The dataset can be accessed from https://console.cloud.google.com/storage/browser/isi-simsurgskill-2021.
\end{abstract}

\section{Introduction}

% problems with current approach... time consuming, subjective
Surgical training requires trainees to master many technical skills before they can move on to performing clinical procedures. The traditional approach for evaluating surgical skills has been through expert supervision~\cite{polavarapu2013100}. However, this requires significant time from expert surgeons and is prone to subjectivity. Hence there is a need for methods that are both more efficient (i.e. faster) and more objective~\cite{maier2018surgical,nagy2017surgical,darzi1999assessing}. Attempts to provide automated skill assessment in the form objective feedback, aim to meet these needs.

% work on CV should help skill assessment
Studies on automated surgical skills assessment~\cite{brown2020bring, zia2018automated, wang2018deep, funke2019video} target many different data modalities, like video, wearable sensors, robot-kinematics, etc. Within the surgical robotics domain, much work has focused on ``video understanding," that is, work that attempts to solve problems like surgical tool detection \cite{bouget2017vision, jin2018tool}, anatomy recognition \cite{madani2022artificial} and task/phase recognition \cite{zia2018surgical, zia2019novel, garrow2021machine, zia2017temporal}. Solving these core computer vision problems can allow for many useful downstream applications, such as automated skills assessment \cite{funke2019video, zia2018video}. 

% OPIs...
Recent research shows that objective performance indicators (OPIs) quantifying tool use and movement during specific tasks within surgeries can differentiate surgeon skill~\cite{chen2018use,brown2020bring,hung2018development}, correlate to outcomes~\cite{hung2019deep,hung2021surgeon}, and describe workflows~\cite{lazar2022commentary}. Paired with advances in computer vision to track surgical tools, OPIs could also be automated. Thus there is the promise of fully automated and objective assessment of surgical skill, though much work remains to be done.

% competitions are good for advancing areas of investigation
Technical competitions, i.e. ``Challenges," have become an integral part of many top computer vision and machine learning conferences. Such challenges have made significant contributions in advancing algorithms that impact surgical assistive technologies. They also serve as a platform making datasets available to the public. MICCAI has hosted many such challenges in the recent past, with Endoscopic Vision (EndoVis) challenge being one  example~\cite{stefanie_speidel_2021_4572973, zia2021surgical, allan2017, allan2018, wagner2021comparative, nwoye2022cholectriplet2021}. The EndoVis challenge is hosted annually at MICCAI and consists of multiple sub-challenges that target various research problems in the surgical domain. To promote and foster work on automated surgical skills assessment, our group hosted the SimSurgSkill 2021 challenge. 

In the SimSurgSkill challenge, competitors used videos of simulated surgical tasks to predict OPIs, which served as a proxy for skill. Specifically, a large dataset consisting of videos of simulated surgical tasks and the resulting OPIs were released to participants. The videos were obtained during virtual reality (VR) robotic training sessions from surgical trainees of varying skill levels. In addition, annotated bounding boxes (for tools and needles) were made available. Teams were tasked with localizing the surgical tools as a first step, and then predicting the resulting OPIs. Overall, we found that the teams did generally well in localizing surgical tools but had difficulty in predicting OPI values. Extensive details regarding the challenge are given in the following sections. 

%We end by proposing future work to further the field.

\section{SimSurgSkill 2021 Challenge Description}
\subsection{Overview}

The SimSurgSkill 2021 challenge took place at the Medical Image Computing and Computer Assisted Interventions (MICCAI) 2021 conference as a sub-challenge of the annual Endoscopic Vision (EndoVis) challenge. The challenge was designed to be based on BIAS \cite{maier2020bias} standards and a full design document is attached (see Appendix~\ref{appendix}). Details regarding the dataset and catagories for this challenge are given below.

\subsection{Data}
\begin{figure}[tb]
\centering
\includegraphics[width= 1\linewidth,clip,trim=8pt 0pt 0pt 0pt]{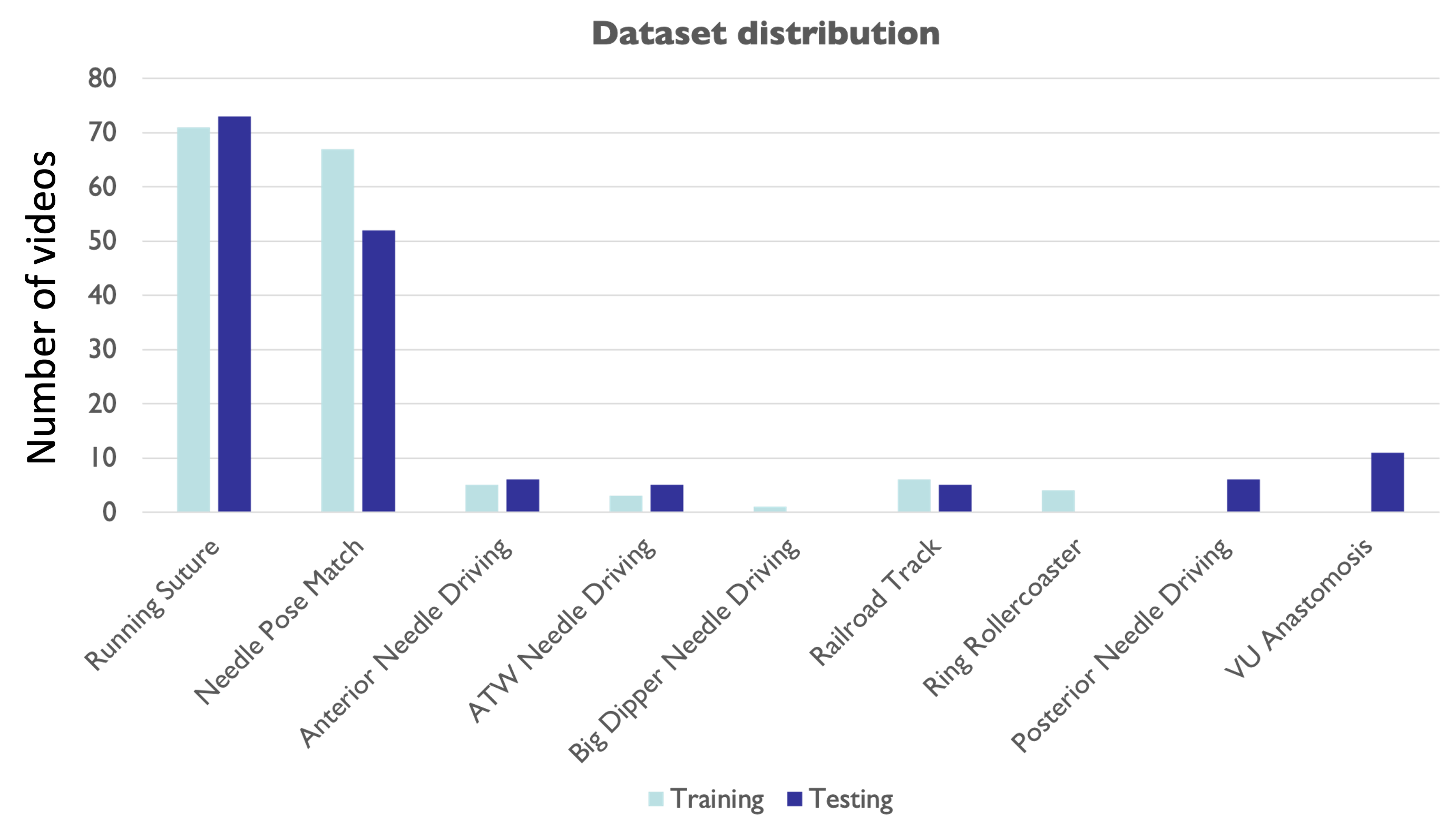}
\caption{Number of video clips across different VR exercises for model training and testing}
\label{dataset_dist}
\end{figure}

% ? is this the right way to refer to the simulator?
%The dataset consisted of virtual reality (VR) videos captured from the simulator environment of the da Vinci robotic system. 
The dataset consisted of videos, video annotations and objective performance indicators (OPIs).
The videos were obtained from multiple surgical virtual reality exercises being performed by participants of varying skill levels. A total of 315 video recordings of various VR exercise were included in the dataset, including 157 videos for model training and 158 videos for testing. As shown in Figure \ref{dataset_dist}, the majority of training and testing cases are from running suture and needle pose matching exercises. To test model generalizibility, posteriror needle driving and VU anatomosis were included only in the testing set. In addition to the videos, bounding boxes of the visible surgical tools and needle were provided, along with OPIs for the tasks performed within each video. 
%In addition to the videos, there are two types of labels that come with the dataset: bounding boxes of the surgical tools and needle, and objective performance indicators (OPIs). 

\subsubsection{Bounding box annotation}
\begin{figure}[tb]
\centering
\includegraphics[width= 1\linewidth,clip,trim=8pt 0pt 0pt 0pt]{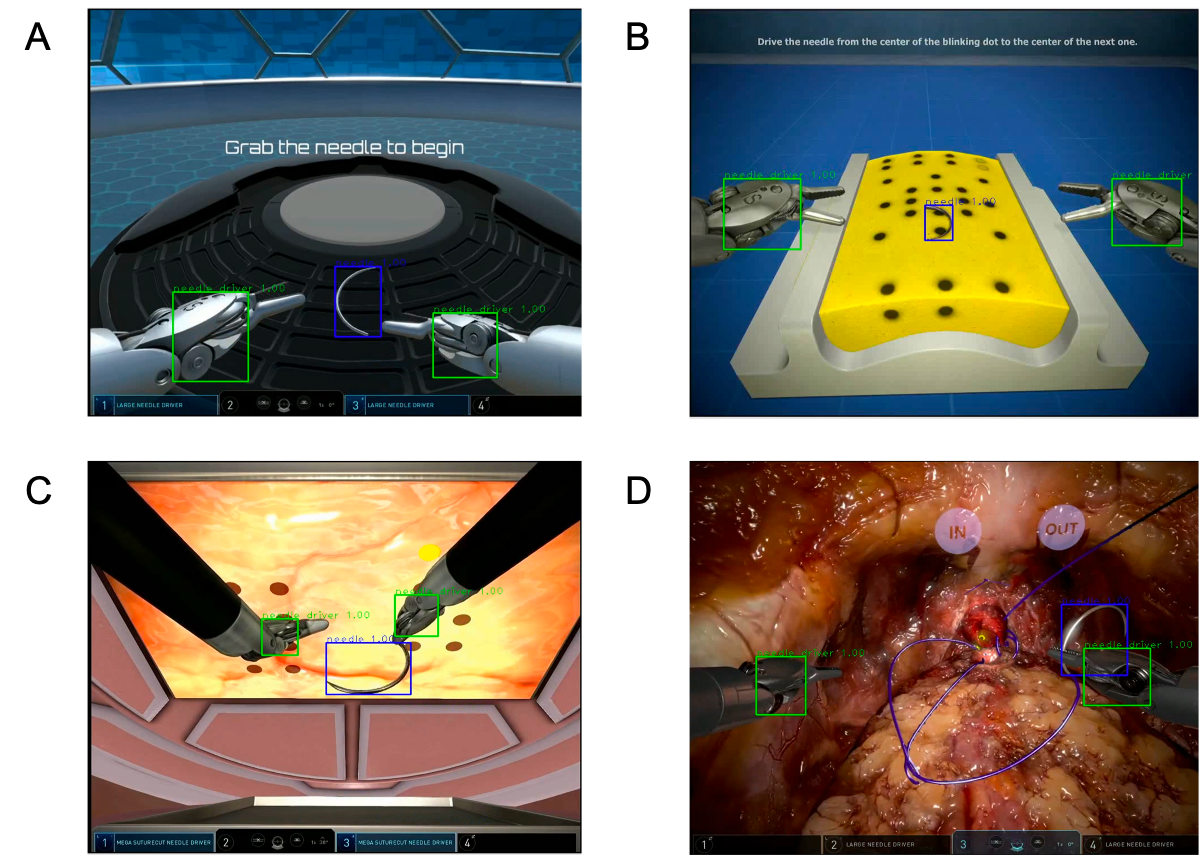}
\caption{Examples of bounding box annotations from training and testing dataset. A) needle pose match exercise. B) running suture. C) anterior needle driving. D) VU anatomosis}
\label{fig:dataset_bbox}
\end{figure}

A portion of the training data is accompanied with bounding boxes around tool center, or clevis, and the needle. The videos are first downsampled to 1 fps and then annotated through crowd-sourcing. A few example annotations and video frames are provided in Figure \ref{fig:dataset_bbox}. The  needle in the needle pose match exercise was not considered of interest and was excluded from annotations. Consequently, false detections of the  needle were considered misclassifications and penalized. 

\subsubsection{Objective performance indicators (OPI)} \label{opi}
For each video, values for 3 OPIs were also provided. These values were generated by the VR engine automatically, hence no human annotation was required. The OPIs along with their definitions are given below:

\begin{itemize}
\item \textbf{Economy of motion (EOM):}
Length of the total path travelled by the clevis of all instruments in use. This distance is calculated in the 3D space and is measured in centimeters. The final value of this metric is a sum of EOM of all tools in use.

\item \textbf{Instrument out-of-view (IOV):}
Count of the total number of times the clevis of a tool goes out of view. Each time any of the instrument's clevis goes out of view, the IOV metric is incremented. The final value of IOV metric is a sum of all IOV events for all tools in use. 

\item \textbf{Needle drop (ND):}
Count of the number of times the needle is dropped.
\end{itemize}

\noindent
The training and testing datasets consists of VR exercises video clips. The VR exercises were completed using the da Vinci simulator by participants of varying skill levels. The videos were captured at 30 fps with a resolution of 720p (1280*720) from one channel of the endoscope.

The dataset has now been released publicly at https://console.cloud.google.com/storage/browser/isi-simsurgskill-2021 but can only be used for non-commercial purposes.

\subsection{Challenge Categories}
\begin{figure}[tb]
\centering
\includegraphics[width= 1\linewidth,clip,trim=10pt 0pt 0pt 0pt]{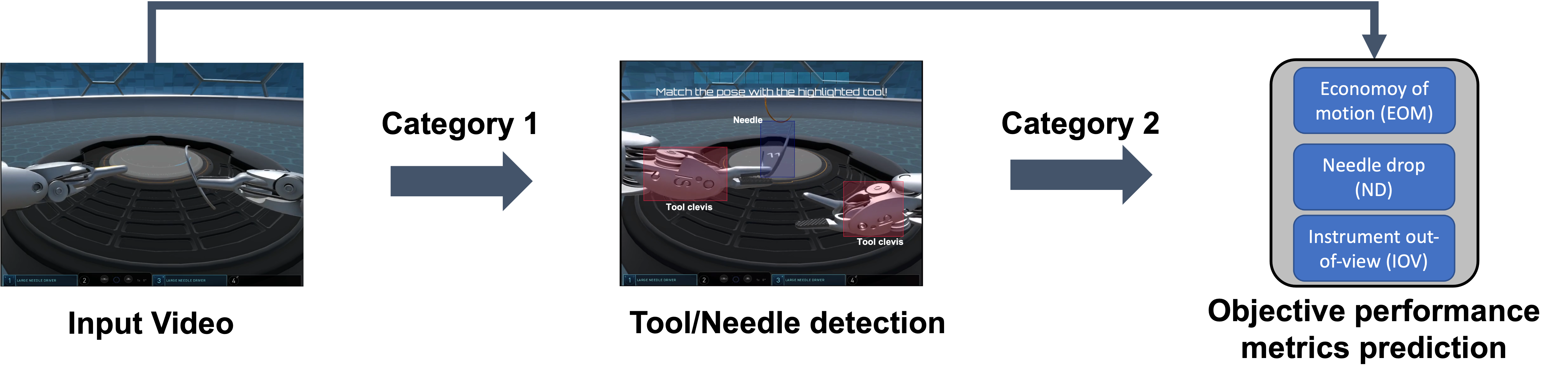}
\caption{The two categories within the challenge.}
\label{categories}
\end{figure}

The challenge was divided into two categories (Figure \ref{categories}). Participating teams could participate individually in either category or both. Details regarding each category follow below.

\subsubsection{Category 1: Surgical tool and needle detection}
In this category, the teams were required to train models that take VR video as input to detect and localize needle drivers and needles using bounding boxes. For this object detection task, evaluation was performed at frame-level at 1 fps. 
Mean average prevision (mAP) of needle and needle drivers were averaged over discrete Intersection over Union (IoU) values (0.5 to 0.95 in increments of 0.05). 
%Mean average prevision (mAP) of needle and needle drivers averaged at different Intersection over Union (IoU) values 0.5:0.05:0.95 was used. 
Note that this evaluation metric is the same as the primary metric used in MS-COCO object detection challenge~\cite{lin2014microsoft}.

\subsubsection{Category 2: Objective performance indicators prediction}
In this category, the teams were required to predict the objective performance indicators using the videos. The three OPIs considered here include needle drop (ND) count, instrument out-of-view (IOV) count, and economy of motion (EOM) with more details listed in section \ref{opi}. Teams could either train models that train directly on video and predict these metrics, or they could use the bounding boxes predicted from category 1 and generate logic to predict the objective metric values. 
To evaluate the performance, mean squared error (MSE) was used for ND and IOV counts, while Pearson correlation coefficient was used for EOM. A ranking table was then produced to rank each team on the three performance metrics. The final ranking of each team was generated by multiplying the 3 rankings. For example, if a team has rankings 1(for ND), 3 (for IOV) and 4 (for EOM), the final ranking will be 12 (1 x 3 x 4). The winner of this category was the one with the highest final rank.

\section{Team Submissions}
There were a total of 16 teams/individuals that had requested access to this challenge's dataset and made an intent to participate. However, only 3 teams were able to make a timely, complete submission to one or both of the categories. Details regarding the methodology of these 3 teams are given below.

\subsection{Muroran Institute of Technology}

The first participating team was from Muroran Institute of Technology, Japan and consisted of only one member - Satoshi Kondo. For Category 1, this team proposed to use EfficientDet \cite{efficientdet} for tool detection. The object detector was pretrained with MS-COCO dataset~\cite{lin2014microsoft} and fine-tuned only the head with the SimSurgSkill dataset. Object detections having scores higher than a threshold of 0.15 were selected until one needle and two surgical tool clevises were found. Since only objects having scores higher than a threshold were selected, the object detector sometimes missed objects. In the case the object detector is unable to find one needle and two surgical tool clevises, this method tried to find the missed object by tracking. The Kernelized Correlation Filter (KCF)~\cite{henriques2014high} was used as the object tracker. The tracker used the most recent object rectangle found by the object detector as its template. When the tracker found an object successfully, the tracking result was used as the object detection result. This tracking method was not used for needles since most of area in a rectangle including a needle was background which made it hard to track the needle.

\begin{figure}[tb]
\centering
\includegraphics[width= 1\linewidth,clip,trim=10pt 0pt 0pt 0pt]{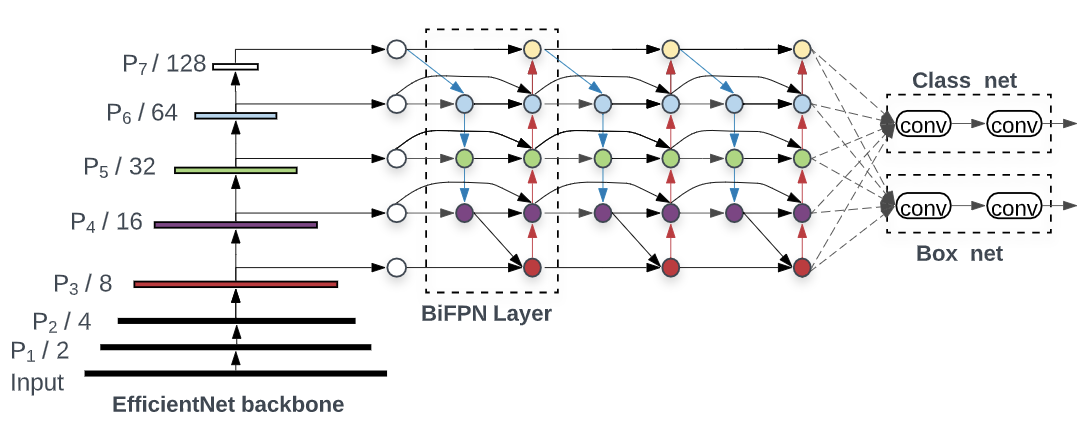}
\caption{Architecture of EfficientDet used by team from Muroran Institute of Technology.}
\label{SK_arch}
\end{figure}

In Category 2, this team used different logic for predicting the three OPIs. For economy of motion (EOM), as the 3D coordinates or camera parameters of the VR simulation system were not known, it was assumed that the size of object is proportional to the distance between the object and the virtual camera. The distance (in pixels) was accumulated along the object trajectory normalized with the object size. The accumulated distance for a whole video was then multiplied by a constant of 1.68 to convert its unit from pixels to 3D volume space.

For instrument out-of-view (IOV), this method utilized the output from the object detection results. When the left or right clevis of a tool was not detected for a duration of 0.5 second, the value of IOV was counted up. For needle drop (ND), there were two conditions defined for counting. The first condition was met when the needle is lower than both surgical tool clevises in image coordinates. The second condition was met when the distance from the needle to two surgical tool clevises was greater than twice the object size. When any of these conditions was satisfied for a duration of 0.5 second, the value of ND was counted up.

\subsection{UCL $+$ Shandong University}
The second team was a collaboration between University College London (UCL) and Shandong University, China and consisted of 3 team members - Jinfan Zhou, Evangelos Mazomenos and Danail Stoyanov. This team only participated in category 2 of the challenge. This team proposed an architecture combining an object detection model and an action recognition model. They first input video frames at 1fps to the object detector to estimate bounding boxes designating the 2D locations of needle and clevises. The sequence of bounding box estimation was used to formulate the Sequence of Interest (SoI) feature which is employed in the detection of ND and IOV events. Then, the SoI was fed at 30fps to both an action recognition model and the previous object detector to produce the final prediction results. Finally, the location information of the bounding boxes was used to estimate the EOM of needle drivers. 

\begin{figure}[tb]
\centering
\includegraphics[width= 1\linewidth,clip,trim=5pt 0pt 5pt 0pt]{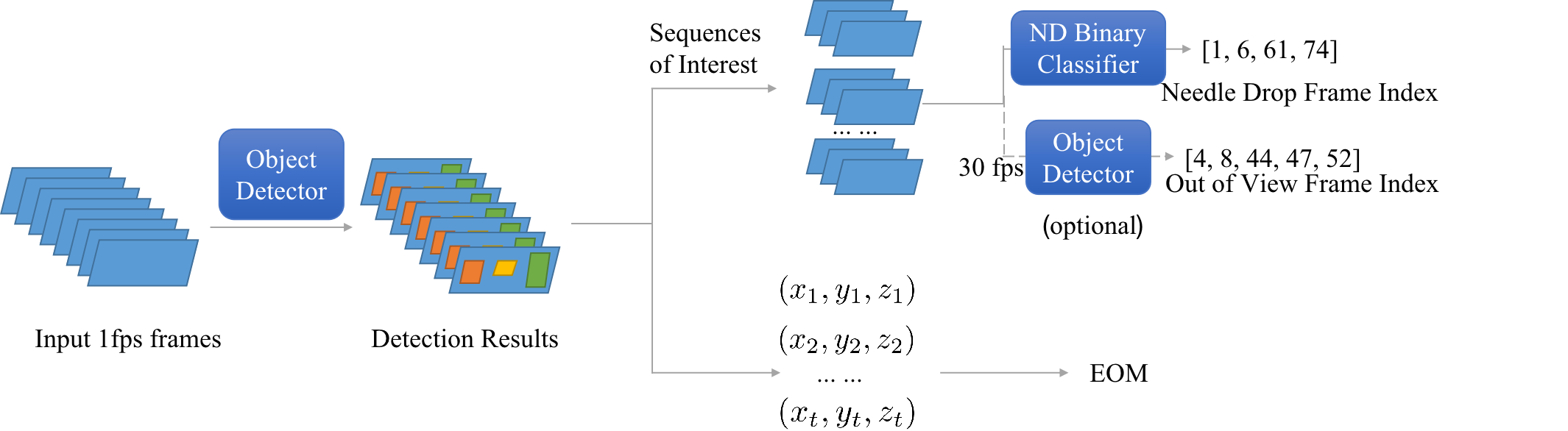}
\caption{UCL $+$ Shandong university's team proposed architecture combining an object detection and an action recognition model.}
\label{jinfan_model_arch}
\end{figure}

For Object Detection, this team trained a Faster R-CNN and Deformable DETR on SimSurgSkill using the open-source detection toolbox MMDetection. ResNet-50 was used as the backbone network for feature extraction, Feature Pyramid Network was used as neck network to fuse the features in different scales, along with a Region Proposal Network to generate region proposals. Fig.~\ref{jinfan_model_arch} shows the architecture of proposed Faster R-CNN model. To train the detector, the team used Stochastic Gradient Descent (SGD) for optimization with a learning rate of 0.02, momentum of 0.9 and weight decay of 1e-4. The training was run for 20 epochs with a batch size of 8. 

\begin{figure}[tb]
\centering
\includegraphics[width= 0.9\linewidth,clip,trim=10pt 0pt 0pt 0pt]{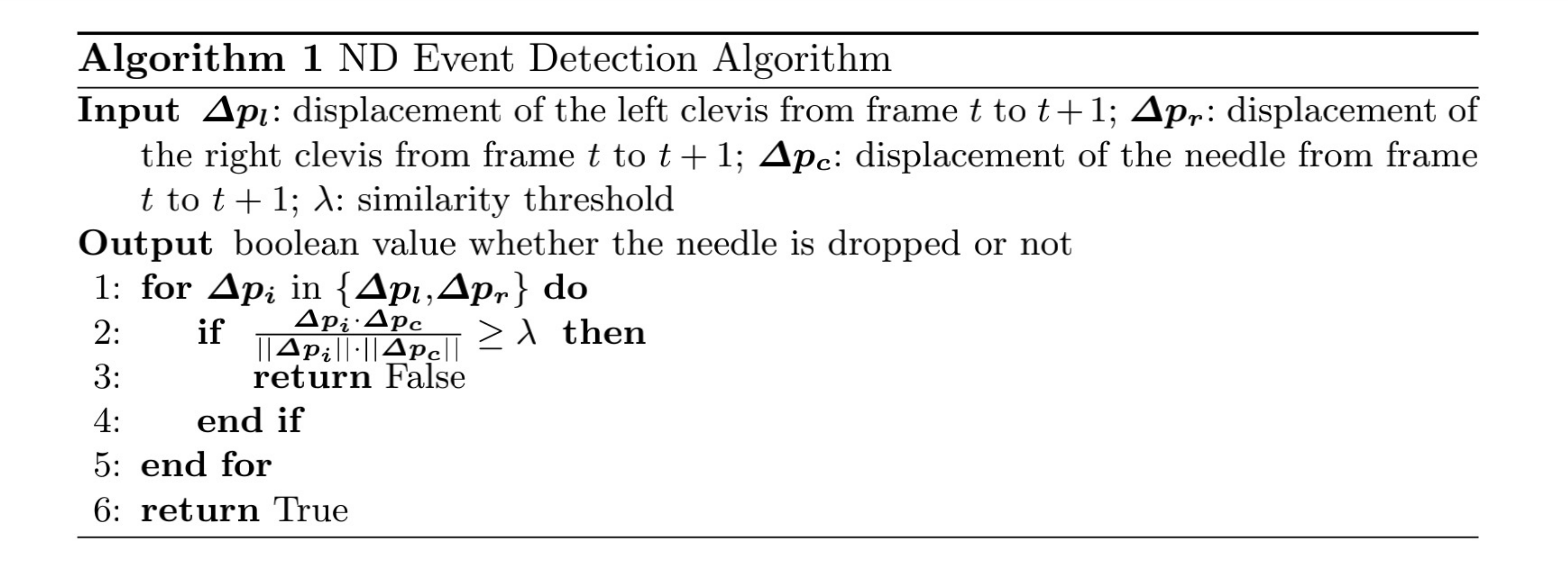}
\caption{UCL $+$ Shandong university's team proposed needle drop detection algorithm. The needle drop was detected by comparing the cosine similarity between the needle and each clevis.}
\label{jinfan_algorithm}
\end{figure}

For Instrument Out-of-View (IOV) detection, the team utilized the output of object detector and calculated a coarse count of IOV events by simply grouping together sequences of frames which contain less than two clevises and calculating their total number. For Needle Drop (ND) Estimation, the team used an Inception-based I3D model for feature extraction and replaced the original prediction head with a two-layer network to provide a binary classification. The selected frames with 63 (32 previous and 31 after) adjacent frames in 30fps were used as input to the I3D model. The training was performed using SGD optimization with an initial learning rate of 0.1, momentum of 0.9, a weight decay of 1e-7 and batch size of 8. The team used a weighted cross entropy loss with a positive(dropped)-to-negative(non-dropped) ratio of 3:1 to avoid over-fitting due to sample imbalance. In addition to the frame-based classifier, the team also proposed an algorithm for ND event detection by comparing the cosine similarity between the needle and each clevis, as shown in Fig.~\ref{jinfan_algorithm}.

% \begin{figure}[tb]
% \centering
% \includegraphics[width= 1\linewidth,clip,trim=20pt 0pt 20pt 0pt]{images/jinhan_needle_drop_method.pdf}
% \caption{Jinfan needle drop estimation.}
% \label{jinhan_needle_drop_classifier}
% \end{figure}

\subsection{UCL}

\begin{figure}[tb]
\centering
\includegraphics[width= 1\linewidth,clip,trim=10pt 0pt 0pt 0pt]{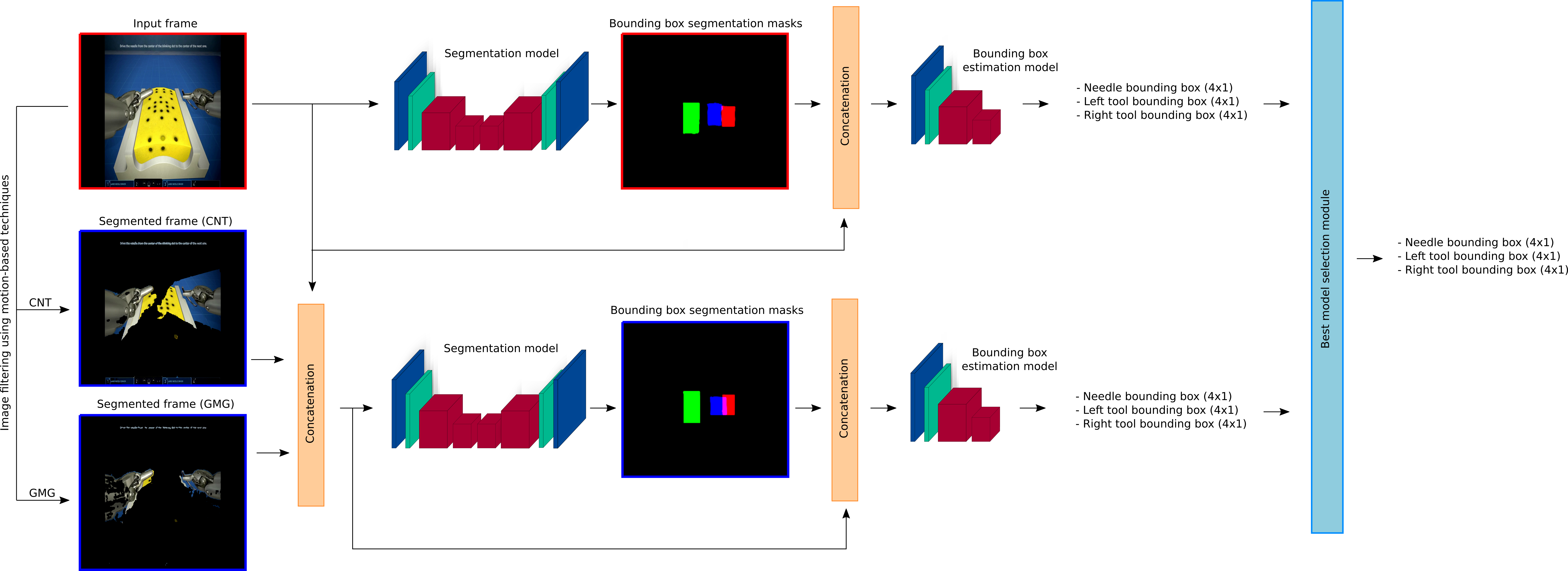}
\caption{Pipeline of UCL team's proposed method. The architecture comprises two twin models, one trained on simple frames while the other was trained on simple frames concatenated with segmented frames. Each model is composed by a segmentation module followed by a bounding box prediction module.}
\label{emanuele_model_arch}
\end{figure}

The last participating team was from University College London and consisted of 3 team members - Emanuele Colleoni, Dimitris Psychogyos and Yueming Jin. This team addressed the tool and needle location tasks by exploiting motion-based video background subtraction as a pre-processing step, along with Mask-RCNN based approaches for object detection. The team hypothesized that by doing so they could reduce the noise introduced by backgrounds not seen during the training phase, thus mitigating over-fitting of their proposed network. Once a bounding box model was trained, the team tackled EOM and IOV tasks based on bounding box predictions. An additional binary classification model was trained to determine whether or not the tool it grabbed was one of the two tools for each frame. Fig.~\ref{emanuele_model_arch} shows the overall workflow of this team's proposed architecture.

Since the background in training videos provided in this challenge does not move continuously, the team proposed to facilitate object detection by first removing the background based on pixel movements over time. Two detection models based on simplified Mask-RCNN were built for tool detection and needle detection using raw image frames and a concatenation of raw and processed (background removal) frames, respectively, as input. Each model consisted of a first step where a segmentation network performs a pixel-based detection of the tool clevis and needle. The output of the segmentation network and the original input were concatenated and then processed into the object detection model. The final outputs were three bounding boxes directly correlated to a specific tool (needle, left/right clevis) for each frame.

Moreover, the team used a Best Model Selection (BMS) module to select the best prediction from the two detection models based on the Intersection over Union (IoU) score between the generated bounding boxes and the segmentation prediction for each frame. The best prediction for each tool was chosen by selecting the model with the highest IoU score, where a tool was considered `not present' when the IoU fell below the threshold $\tau=0.05$ in both models. Both segmentation and bounding box models used Imagenet pre-trained Xception net as backbone. 

For needle drop detection, the team trained an image-based binary classification model to detect whether or not one of the two tools is grabbing the needle. As the team experimentally observed that training such a model on full frames produced unreliable results, they produced a new dataset by first extracting the needle from all frames using the bounding boxes predicted at the previous stage and then resizing the crops to 128$\times$128. The same Xception net was used to build the binary classification model as backbone.

\section{Results}

\subsection{Category 1: Surgical tool and needle detection}
\begin{figure}[h]
\centering
\includegraphics[width= 1\linewidth,clip,trim=10pt 0pt 0pt 0pt]{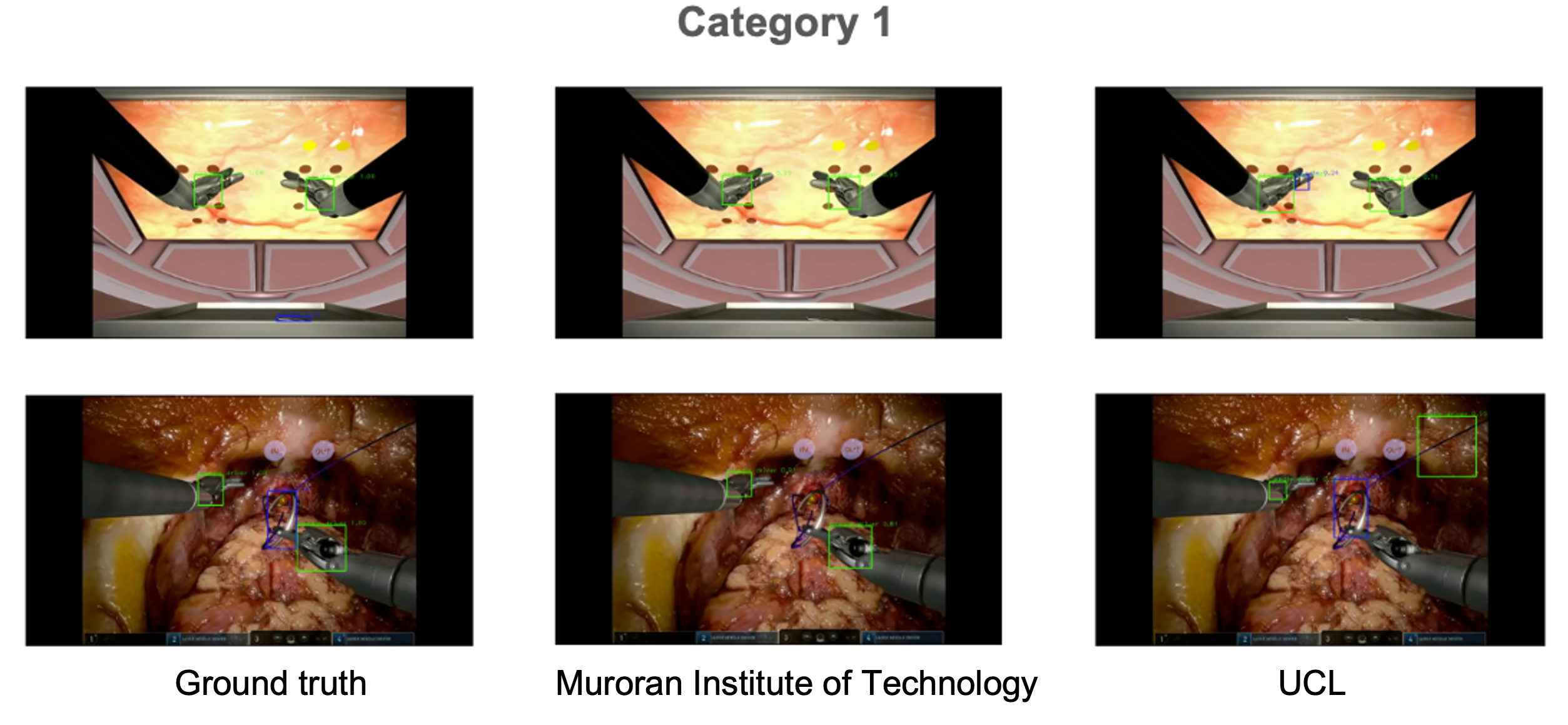}
\caption{Example bounding box detections on the test set.}
\label{fig:bbox_cat_1}
\end{figure}

Two teams submitted to Category 1. Example bounding box predictions on two VR exercises from each team in comparison to ground truth labels are shown in Figure \ref{fig:bbox_cat_1}. In addition to the primary evaluation metric $(AP_{0.5:0.95})$ used to decide winning teams, additional metrics are also provided in Table \ref{tab:cat_1_table} for comparison.

% \begin{figure}[h]
% \centering
% \includegraphics[width= 1\linewidth,clip,trim=10pt 0pt 0pt 0pt]{images/results_category_1.png}
% \caption{Results from Category 1: Surgical tool and needle detection}
% \label{fig:results_cat_1}
% \end{figure}

\begin{table}
\centering
\begin{tabular}{ |c|c|c|c|c| }
 \hline
 \multicolumn{5}{|c|}{Category 1} \\
 \hline
 Team & $AP_{0.5:0.95}$ & $AP_{0.5}$ & $AP_{0.75}$ & $AR_{0.5:0.95}$ \\ 
 \hline
 UCL & 0.33 & \textbf{0.71} & 0.26 & 0.37  \\  
 \hline
 Muroran Institute of Technology & \textbf{0.35} & 0.65 & \textbf{0.34} & \textbf{0.44} \\  
 \hline
\end{tabular}
\caption{\label{tab:cat_1_table} Category 1 objection detection metrics}
\end{table}

\subsection{Category 2: Objective performance indicators prediction}

\begin{figure}[t!]
\centering
\includegraphics[width= 1\linewidth,clip,trim=10pt 0pt 0pt 0pt]{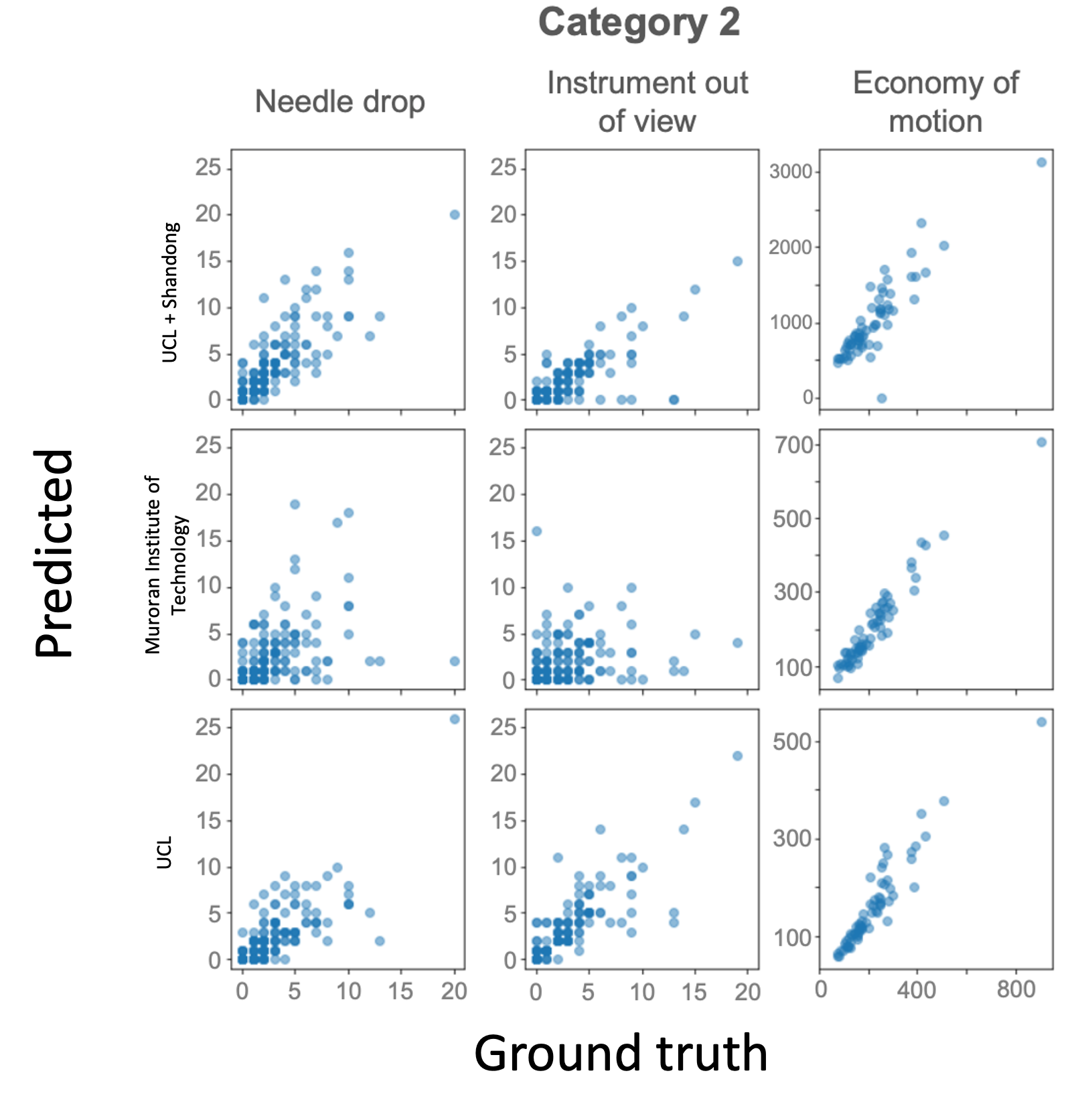}
\caption{Results from Category 2: Objective performance indicators prediction}
\label{fig:results_cat_2}
\end{figure}

All 3 teams prepared submissions for Category 2. Since Needle drop and Instrument out-of-view counts are integer values, we compared ground truth with predictions using mean-squared error. However, we used the Pearson correlation coefficient to compare the Economy of motion ground truth and predicted values. Figure \ref{fig:results_cat_2} shows some scatter plots comparing ground truth with predicted values, whereas Table \ref{tab:cat_2_table} shows the evaluation metric values 

\begin{table}
\centering
\begin{tabular}{ |c|c|c|c| }
 \hline
 \multicolumn{4}{|c|}{Category 2} \\
 \hline
 Team & ND (mse) & IOV (mse) & EOM (pearsonr) \\ 
 \hline\hline
 UCL & \textbf{4.413} & \textbf{3.796} & 0.942 \\  
 \hline
 Muroran Institure of Technology & 11.036 & 12.778 & \textbf{0.964} \\  
 \hline
 UCL $+$ Shandong & 5.167 & 5.144 & 0.879 \\  
 \hline
\end{tabular}
\caption{\label{tab:cat_2_table} Evaluation metric values for predictions of Needle drop (ND), Instrument out-of-view (IOV), and Economy of motion (EOM). Predicted Needle drop and Instrument out of view counts were compared to ground truth using mean-squared error (mse) while Economy of motion predictions were compared to ground truth values using the Pearson correlation coefficient.}
\end{table}

\section{Discussions}
In Category 1, teams had similar performance in tool and needle detection with team Muroran Institure of Technology marginally out-performing Team UCL in all metrics except for one. Though the teams used different approaches, it is unclear whether their similarity in performance indicates a performance limit on the data set. 

In Category 2, teams predicted the EOM with high correlation to the real values with all Pearson's r coefficients from teams $>$0.87. However, predicted values for EOM were quite different from the real values indicating that models performed well at differentiating between relative motion between videos but did not correspond directly the distance measured by the simulation. Furthermore, the high correlation of predicted values with actual EOM across all teams suggests that different modeling approaches are equally capable of estimating tool motion.

The performance of teams in predicting ND and IOV was more variable. Team UCL had the lowest mean-squared-error in the prediction of ND and IOV while team Muroran Institute of Technology had the highest error in prediction. Team UCL $+$ Shandong marginally under-performed team UCL. These differences in performance between team UCL and team Muroran Institute of Technology in predicting ND and IOV while similarity in performance in tool detection (Category 1) suggests that certain methods of processing tool and needle detection to arrive at measure of needle drops or instruments-out-of-view are better than others.  

Overall, from all teams, we see a step-wise approach of estimating objective performance indicators by layering models. We also start to see how model performances in Category 1 interact to influence the performance of the teams in Category 2. This provides an interesting example of how complex problems like automated objective assessment of surgical skill may require an orchestration of multiple models which directly or indirectly influence one another. The complex technical context of an orchestration of models for skill assessment must also be explainable and actionable. In this challenge, we target 3 easily understood OPIs for which higher values indicate lower skill. However, there may be future use-cases where skill is inferred from combining multiple OPIs. With these considerations, we identify a focus on interpretability as an important and valuable future direction in automated surgical skill assessment.

\section{Conclusion}
In this challenge, we asked the surgical data science community to develop computer vision algorithms which predict 3 objective measures of surgical skill related to instrument-use and needle-handling from videos of surgical tasks completed in a simulated environment. The objective performance indicators used in this challenge were quite interpretable and, generally, have a simple relationship with skill. However, since we did not have skill labels for the dataset, it is unclear whether and how the inference of skill from OPIs would be influenced by the error in estimating the OPIs from the different approaches. 

Nonetheless, initial results from the teams suggest that the target of estimating OPIs from computer vision algorithms seems reachable. Correlation to EOM---estimation of tool motion---is quite high for all approaches. For IOV and ND, specific approaches are able to perform quite well but other approaches are less successful. Approaches layered models to estimate OPIs suggesting that the performances in Category 2 were influenced by the performance in Category 1. We hope that the dataset and approaches presented here enable and encourage further investigations into automated surgical skill assessment---an important innovation which could accelerate and scale surgical training.

\bibliographystyle{unsrt}
\bibliography{references}

\newpage

\appendix
\section{Appendix}
\label{appendix}
The initial design document for the challenge can be found below.


\section*{Supplementary material}

# Suppl 4: Structured description of a challenge design

## SUMMARY

**Item 1: Title**
a) Use the title to convey the essential information on the **challenge mission**.

b) Preferable, provide a short **acronym** of the challenge (if any).

**Item 2: Abstract**
Provide a **summary** of the challenge purpose. This should include a general introduction in the topic from both a biomedical as well as from a technical point of view and clearly state the envisioned technical and/or biomedical impact of the challenge.

**Item 3: Keywords**
List the primary **keywords** that characterize the challenge.

## CHALLENGE ORGANIZATION

**Item 4: Organizers**
a) Provide information on the **organizing team** (names and affiliations).

b) Provide information on the **primary contact person**.

**Item 5: Lifecycle type**
Define the intended **submission cycle** of the challenge. Include information on whether/how the challenge will be continued after the challenge has taken place.
*Examples:*
- One-time event with fixed submission deadline
- Open call
- Repeated event with annual fixed submission deadline

**Item 6: Challenge venue and platform**
a) Report the **event** (e.g. conference) that is **associated** with the challenge (if any).

b) Report the **platform** (e.g. grand-challenge.org) used to run the challenge.

c) Provide the **URL** for the challenge website (if any).

**Item 7: Participation policies**
a) Define the **allowed user interaction** of the algorithms assessed (e.g. only (semi-) automatic methods allowed).

b) Define the policy on the **usage of training data**. The data used to train algorithms may, for example, be restricted to the data provided by the challenge or to publicly available data including (open) pre-trained nets.

c) Define the **participation policy for members of the organizers' institutes**. For example, members of the organizers' institutes may participate in the challenge but are not eligible for awards.

d) Define the **award policy**. In particular, provide details with respect to challenge prizes.

e) Define the policy for **result announcement**.
   *Examples:*
   - Top three performing methods will be announced publicly.
   - Participating teams can choose whether the performance results will be made public.

f) Define the **publication policy**. In particular, provide details on ...
   - … who of the participating teams/the participating teams' members qualifies as author
   - … whether the participating teams may publish their own results separately, and (if so)
   - … whether an embargo time is defined (so that challenge organizers can publish a challenge paper first).

**Suppl 4:** Structured description of a challenge design

**Item 8: Submission method**
a) Describe the method used for result submission. Preferably, provide a link to the **submission instructions**.
*Examples:*
- Docker container on the Synapse platform. Link to submission instructions: <URL>
- Algorithm output was sent to organizers via e-mail. Submission instructions were sent by e-mail.

b) Provide information on the possibility for participating teams to **evaluate** their **algorithms before submitting** final results. For example, many challenges allow submission of multiple results, and only the last run is officially counted to compute challenge results.

**Item 9: Challenge schedule**
Provide a **timetable** for the challenge. Preferably, this should include
- the release date(s) of the training cases (if any)
- the registration date/period
- the release date(s) of the test cases and validation cases (if any)
- the submission date(s)
- associated workshop days (if any)
- the release date(s) of the results

**Item 10: Ethics approval**
Indicate whether **ethics approval** is necessary for the data. If yes, provide details on the ethics approval, preferably institutional review board, location, date and number of the ethics approval (if applicable). Add the URL or a **reference to the document** of the ethics approval (if available).

**Item 11: Data usage agreement**
Clarify how the data can be used and distributed by the teams that participate in the challenge and by others during and after the challenge. This should include the explicit **listing of the license** applied.
*Examples:*
- CC BY (Attribution)
- CC BY-SA (Attribution-ShareAlike)
- CC BY-ND (Attribution-NoDerivs)
- CC BY-NC (Attribution-NonCommercial)
- CC BY-NC-SA (Attribution-NonCommercial-ShareAlike)
- CC BY-NC-ND (Attribution-NonCommercial-NoDerivs)

**Suppl 4:** Structured description of a challenge design

**Item 12: Code availability**
a) Provide information on the **accessibility of the organizers' evaluation software** (e.g. code to produce rankings). Preferably, provide a link to the code and add information on the supported platforms.

b) In an analogous manner, provide information on the **accessibility of the participating teams' code.**

**Item 13: Conflicts of interest**
Provide information related to conflicts of interest. In particular provide information related to **sponsoring/ funding** of the challenge. Also, state explicitly who had/will have **access to the test case labels** and when.

## MISSION OF THE CHALLENGE

**Item 14: Field(s) of application**
State the **main field(s) of application** that the participating algorithms target.
*Examples:*
- Diagnosis
- Education
- Intervention assistance
- Intervention follow-up
- Intervention planning
- Prognosis
- Research
- Screening
- Training
- Cross-phase

**Item 15: Task category(ies)**
State the **task category(ies)**.
*Examples:*
- Classification
- Detection
- Localization
- Modeling
- Prediction
- Reconstruction
- Registration
- Retrieval
- Segmentation
- Tracking

**Suppl 4:** Structured description of a challenge design

**Item 16: Cohorts**
We distinguish between the *target cohort* and the *challenge cohort*. For example, a challenge could be designed around the task of medical instrument tracking in robotic kidney surgery. While the challenge could be based on *ex vivo* data obtained from a laparoscopic training environment with porcine organs (challenge cohort), the final biomedical application (i.e. robotic kidney surgery) would be targeted on real patients with certain characteristics defined by inclusion criteria such as restrictions regarding gender or age (target cohort).

a) Describe the **target cohort**, i.e. the subjects/objects from whom/which the data would be acquired in the final biomedical application.

b) Describe the **challenge cohort**, i.e. the subject(s)/object(s) from whom/which the challenge data was acquired.

**Item 17: Imaging modality(ies)**
Specify the **imaging technique(s)** applied in the challenge.

**Item 18: Context information**
Provide additional **information given along with the images**. The information may correspond ...
a) ... directly to the **image data** (e.g. tumor volume).

b) ... to the **patient** in general (e.g. gender, medical history).

**Item 19: Target entity(ies)**
a) Describe the **data origin**, i.e. the region(s)/part(s) of subject(s)/object(s) from whom/which the image data would be acquired in the final biomedical application (e.g. brain shown in computed tomography (CT) data, abdomen shown in laparoscopic video data, operating room shown in video data, thorax shown in fluoroscopy video). If necessary, differentiate between target and challenge cohort.

**Suppl 4:** Structured description of a challenge design

b) Describe the **algorithm target**, i.e. the structure(s)/subject(s)/object(s)/component(s) that the participating algorithms have been designed to focus on (e.g. tumor in the brain, tip of a medical instrument, nurse in an operating theater, catheter in a fluoroscopy scan). If necessary, differentiate between target and challenge cohort.

**Item 20: Assessment aim(s)**
Identify the **property(ies) of the algorithms to be optimized** to perform well in the challenge. If multiple properties are assessed, prioritize them (if appropriate). The properties should then be reflected in the metrics applied (parameter 26), and the priorities should be reflected in the ranking when combining multiple metrics that assess different properties.
- *Example 1:* Find liver segmentation algorithm for CT images that processes CT images of a certain size in less than a minute on a certain hardware with an error that reflects inter-rater variability of experts.
- *Example 2:* Find lung tumor detection algorithm with high sensitivity and specificity for mammography images.

Corresponding metrics are listed below (parameter 26).

## CHALLENGE DATA SETS

**Item 21: Data source(s)**
a) Specify the **device(s)** used to acquire the challenge data. This includes details on the device(s) used to acquire the imaging data (e.g. manufacturer) as well as information on additional devices used for performance assessment (e.g. tracking system used in a surgical setting).

b) Describe relevant details on the imaging process/**data acquisition** for each acquisition device (e.g. image acquisition protocol(s)).

c) Specify the **center(s)/institute(s)** in which the data was acquired and/or the **data providing platform/source** (e.g. previous challenge). If this information is not provided (e.g. for anonymization reasons), specify why.

d) Describe relevant **characteristics** (e.g. level of expertise) **of the subjects** (e.g. surgeon)/objects (e.g. robot) involved in the data acquisition process (if any).

**Item 22: Training and test case characteristics**
a) State what is meant by one **case** in this challenge. A case encompasses all data that is processed to produce one result that is compared to the corresponding reference result (i.e. the desired algorithm output).
   *Examples:*
   - Training and test cases both represent a CT image of a human brain. Training cases have a weak annotation (tumor present or not and tumor volume (if any)) while the test cases are annotated with the tumor contour (if any).
   - A case refers to all information that is available for one particular patient in a specific study. This information always includes the image information as specified in *data source(s)* (parameter 21) and may include context information (parameter 18). Both training and test cases are annotated with survival (binary) 5 years after (first) image was taken.

b) State the **total number** of training, validation and test cases.

c) Explain **why a total number** of cases and **the specific proportion** of training, validation and test cases was chosen.

d) Mention **further important characteristics** of the training, validation and test cases (e.g. class distribution in classification tasks chosen according to real-world distribution vs. equal class distribution) and justify the choice.

**Item 23: Annotation characteristics**
a) Describe the **method for determining the reference annotation**, i.e. the desired algorithm output. Provide the information separately for the training, validation and test cases if necessary. Possible methods include *manual image annotation*, *in silico ground truth generation* and *annotation by automatic methods*.

If human annotation was involved, state the **number of annotators**.

b) Provide the **instructions given to the annotators** (if any) prior to the annotation. This may include description of a training phase with the software. Provide the information separately for the training, validation and test cases if necessary. Preferably, provide a link to the **annotation protocol**.

c) Provide **details on the subject(s)/algorithm(s) that annotated** the cases (e.g. information on **level of expertise** such as number of years of professional experience, medically-trained or not). Provide the information separately for the training, validation and test cases if necessary.

d) Describe the **method(s) used to merge multiple annotations** for one case (if any). Provide the information separately for the training, validation and test cases if necessary.

**Item 24: Data pre-processing method(s)**
Describe the **method(s) used for pre-processing** the raw training data before it is provided to the participating teams. Provide the information separately for the training, validation and test cases if necessary.

**Item 25: Sources of error**
a) Describe the most relevant **possible error sources related to the image annotation**. If possible, **estimate the magnitude** (range) of these errors, using inter-and intra-annotator variability, for example. Provide the information separately for the training, validation and test cases, if necessary.

b) In an analogous manner, describe and quantify **other relevant sources of error**.

## ASSESSMENT METHODS

**Item 26: Metric(s)**
a) Define the **metric(s) to assess a property of an algorithm**. These metrics should reflect the desired algorithm properties described in *assessment aim(s)* (parameter 20). State which metric(s) were used to compute the ranking(s) (if any).
- *Example 1:* Dice Similarity Coefficient (DSC) and run-time
- *Example 2:* Area under curve (AUC)

b) **Justify why** the metric(s) was/were chosen, preferably with reference to the biomedical application.

**Suppl 4:** Structured description of a challenge design

**Item 27: Ranking method(s)**
a) Describe the **method used to compute a performance rank** for all submitted algorithms based on the generated metric results on the test cases. Typically the text will describe how results obtained per case and metric are aggregated to arrive at a final score/ranking.


b) Describe the method(s) used to manage **submissions with missing results** on test cases.


c) **Justify why** the described ranking scheme(s) was/were used.


**Item 28: Statistical analyses**
a) Provide **details for the statistical methods** used in the scope of the challenge analysis. This may include
- description of the **missing data handling**,
- details about the assessment of **variability of rankings**,
- description of any method used to assess **whether the data met the assumptions**, required for the particular statistical approach, or
- indication of any **software product** that was used for all data analysis methods.


b) **Justify why** the described statistical method(s) was/were used.


**Item 29: Further analyses**
Present further analyses to be performed (if applicable), e.g. related to
- **combining algorithms** via ensembling,
- **inter-algorithm variability**,
- **common problems/biases** of the submitted methods, or
- **ranking variability**.



\end{document}